\documentclass[11pt]{article}

\usepackage[final]{acl}
\usepackage{times}
\usepackage{latexsym}
\usepackage{enumitem}
\usepackage[T1]{fontenc}

\usepackage[utf8]{inputenc}

\usepackage{microtype}

\usepackage{inconsolata}

\usepackage{graphicx}
\usepackage{booktabs}
\usepackage{multirow}
\usepackage[most]{tcolorbox}
\usepackage{CJKutf8}
\newcommand{\zh}[1]{\begin{CJK*}{UTF8}{gbsn}#1\end{CJK*}}
\newcommand{\ja}[1]{\begin{CJK*}{UTF8}{min}#1\end{CJK*}}
\newcommand{\ko}[1]{\begin{CJK*}{UTF8}{mj}#1\end{CJK*}}

\title{Beyond Monolingual Deep Research: Evaluating Agents and Retrievers with Cross-Lingual BrowseComp-Plus}

\author{
  Yuheng Lu$^{1}$\thanks{Equal contribution.}, Qingcheng Zeng$^{2}$\footnotemark[1], Heli Qi$^{1,3}$, Puxuan Yu$^{4}$, \textbf{Fuheng Zhao}$^{5}$,  \\\textbf{Rui Yang}$^{6}$, \textbf{Hitomi Yanaka}$^{7,3}$, \textbf{Naoto Yokoya}$^{7,3}$, \textbf{Weihao Xuan}$^{7,3}$\thanks{Corresponding author.}  \\
  $^{1}$Waseda University, $^{2}$Northwestern University, $^{3}$RIKEN AIP, $^{4}$Snowflake Inc., \\$^{5}$University of Utah, $^{6}$Duke-NUS Medical School, $^{7}$The University of Tokyo\\
}

\begin{document}
\maketitle

\begin{abstract}
Deep research agents are increasingly evaluated on their ability to search for evidence, reason over retrieved sources, and produce grounded answers. Existing browsing benchmarks, however, largely assume that the user's query and the supporting evidence are written in the same language, leaving open whether agentic search systems can operate when relevant evidence appears in another language. We introduce \textsc{XBCP} (Cross-lingual BrowseComp-Plus), a controlled benchmark that preserves the English question-and-answer space of BrowseComp-Plus but varies the languages of the supporting documents. \textsc{XBCP} instantiates two complementary settings. In the cross-lingual setting, each query is paired with evidence in a single assigned language. In the multilingual setting, the full evidence corpus is distributed equally and randomly across 12 languages spanning high-resource and low-resource regimes. We evaluate four deep research agents using sparse and dense multilingual retrievers, measuring answer accuracy, evidence recall, search behavior, calibration, citation fidelity, and oracle retrieval. Results reveal substantial degradation when evidence is translated. Even strong, dense retrievers lose evidence recall, and agents become less calibrated and cite evidence less reliably. Notably, accuracy remains lower even when all gold evidence is supplied directly. These findings suggest that cross-lingual deep research exposes both retrieval failures and an independent, agent-side difficulty in integrating language-mismatched evidence.
\end{abstract}

\section{Introduction}
\begin{figure*}[t]
\centering
\includegraphics[width=\textwidth]{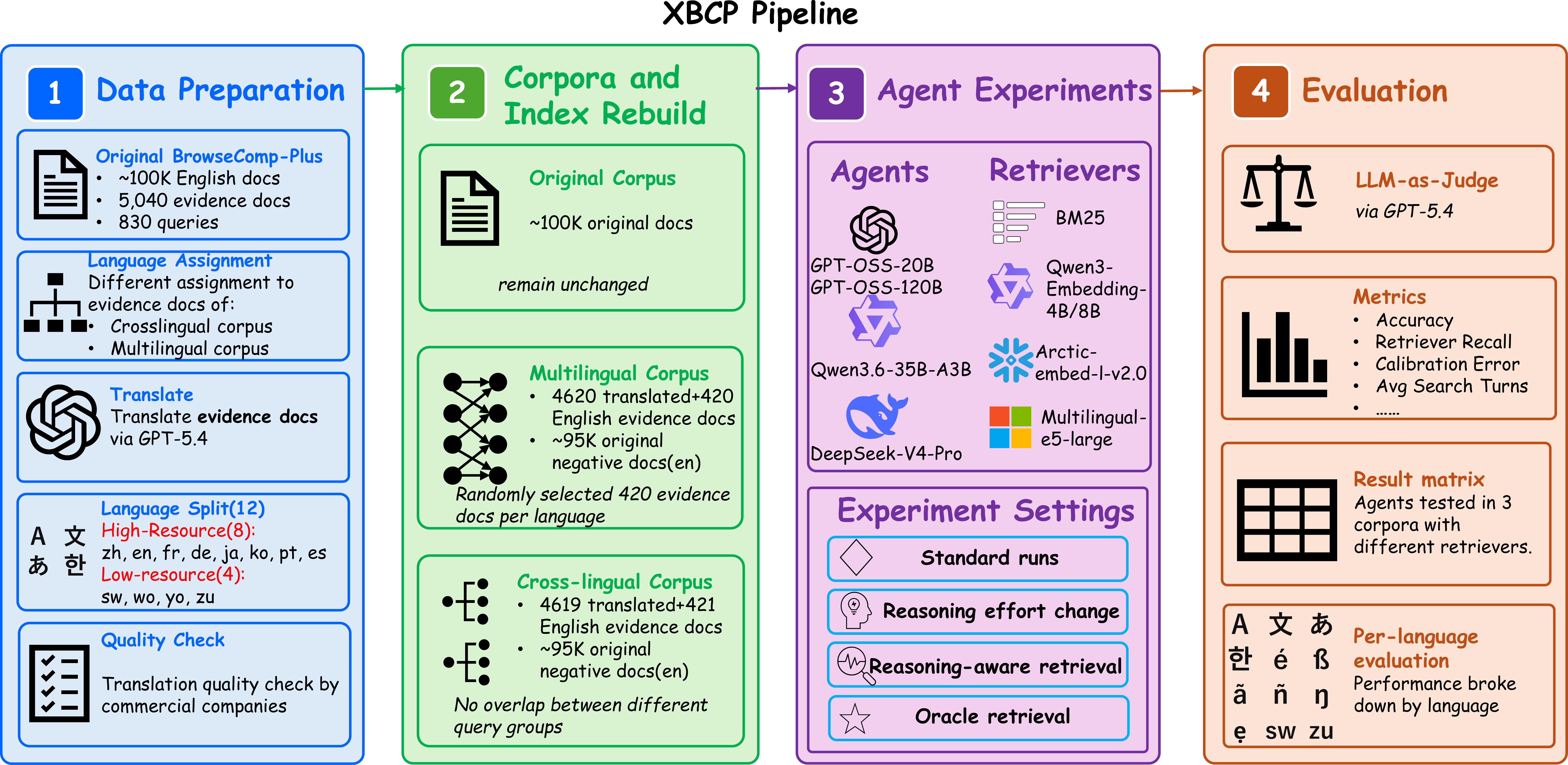}
\caption{\looseness=-1 Overview of the \textsc{XBCP} pipeline. We translate and reorganize the evidence side of BrowseComp-Plus into cross-lingual and multilingual corpora, rebuild retrieval indexes for controlled agent experiments, and evaluate agents and retrievers with end-to-end accuracy, evidence recall, calibration, oracle retrieval, and per-language analysis.}
\label{fig:xbcp-pipeline}
\end{figure*}

\enlargethispage{\baselineskip}
Large language model (LLM) agents represent a shift from models that answer from parametric knowledge alone to systems that actively acquire, filter, and synthesize external evidence. Deep research systems are a representative instance of this shift: given a complex information need, an agent must plan searches, inspect retrieved sources, judge whether the evidence is sufficient, and compose a grounded answer \citep{openai2025deepresearch}. This broader movement has made browsing-based evaluation a central test of agentic capability. BrowseComp \citep{wei2025browsecompsimplechallengingbenchmark} crystallizes the challenge by posing difficult but verifiable questions whose answers require nontrivial web exploration, thereby stressing both search behavior and evidence-grounded reasoning. However, evaluations over live web search measure an entire time-varying system at once, entangling the language model, retrieval method, ranking API, and underlying corpus. BrowseComp-Plus \citep{chen2025browsecompplusfairtransparentevaluation} addresses this limitation by grounding BrowseComp-style questions in a fixed, human-verified corpus with supporting documents and hard negatives, turning browsing evaluation into a controlled setting where retrievers and LLM agents can be studied both separately and in interaction.
\enlargethispage{2\baselineskip}
This controlled view of deep research, however, remains largely confined to monolingual settings. The limitation matters because multilingual and cross-lingual retrieval have long been central concerns in information retrieval, and recent multilingual embedding models have greatly expanded the ability to retrieve across languages \citep{yu2024arcticembed20multilingualretrieval,zhang-etal-2024-mgte,zhang2025qwen3embeddingadvancingtext}. Most evaluations of these models still treat retrieval as a standalone ranking problem: a query is matched against a fixed collection, and success is measured by document-level relevance. This abstraction is useful for isolating retrieval quality, but it does not capture what happens when retrieval is part of an agentic search process. In that setting, the system must issue and refine searches, compare partial evidence, and decide how retrieved information should support an answer. Recent browsing-agent benchmarks beyond English, such as BrowseComp-ZH \citep{zhou2025browsecompzhbenchmarkingwebbrowsing}, broaden the linguistic scope of agent evaluation but remain primarily monolingual: questions, evidence, and answers all stay within the same language. They therefore leave open the genuinely cross-lingual case, where an information need expressed in one language must be answered using evidence written in another. A cross-lingual extension of BrowseComp-Plus is needed to make this setting measurable. Such a benchmark would test whether multilingual retrievers can surface the right evidence during agentic search and whether LLM agents can integrate language-mismatched evidence into faithful answers. 
\enlargethispage{\baselineskip}
To make this setting measurable, we introduce Cross-lingual BrowseComp-Plus (XBCP). To the best of our knowledge, XBCP is the first benchmark to formalize cross-lingual deep research, extending the controlled evaluation paradigm of BrowseComp-Plus from monolingual to multilingual retrieval. XBCP preserves the task structure of BrowseComp-Plus: questions are posed in English, answers are expected in English, and the evidence is grounded in a fixed corpus. The key difference is that the supporting evidence is no longer assumed to be written in the same language as the question. We instantiate this design with two complementary configurations. In the cross-lingual setting, all supporting documents for a given query appear in the same language, while the assigned language varies across queries. This tests whether systems remain robust as otherwise comparable tasks move across languages. In the multilingual setting, the evidence corpus is randomly but equally assigned to 12 languages spanning high-resource and low-resource regimes, enabling controlled evaluation of English queries against language-specific evidence documents. Together, these configurations allow XBCP to evaluate both whether multilingual retrievers can surface language-mismatched evidence during agentic search and whether LLM agents can integrate such evidence into faithful English answers. Our experiments reveal large drops in accuracy and evidence recall across retrievers, reduced citation reliability, and persistent degradation even under oracle retrieval. These findings indicate that cross-lingual deep research stresses both retrieval and agent-side evidence integration. Figure~\ref{fig:xbcp-pipeline} summarizes the construction and evaluation pipeline. Code and datasets are publicly available in our \href{https://github.com/paddler2022/XBCP}{GitHub repository} and \href{https://huggingface.co/datasets/UTokyo-Yokoya-Lab/XBCP/tree/main}{Hugging Face collections}.

\section{Related Works}
\paragraph{Deep Research Systems.} Deep research systems extend tool-augmented LLMs from single-step retrieval to long-horizon information seeking, where agents must plan searches, interact with external sources, verify intermediate evidence, and synthesize grounded answers. OpenAI Deep Research \citep{openai2025deepresearch} exemplifies this paradigm and has motivated a growing line of open research agents that scale the underlying capabilities in different ways: Tongyi DeepResearch \citep{tongyideepresearchteam2026tongyideepresearchtechnicalreport} combines agentic mid-training and post-training with large-scale synthetic trajectories, MiroThinker \citep{miromindteam2026mirothinkerpushingperformanceboundaries} studies model, context, and interaction scaling, and Marco DeepResearch \citep{zhu2026marcodeepresearchunlockingefficient} emphasizes verification-centric training and inference to reduce error propagation in long-horizon search. Benchmarking has also moved toward more demanding settings, including Chinese web browsing in BrowseComp-ZH \citep{zhou2025browsecompzhbenchmarkingwebbrowsing}, expert-level financial search in FinSearchComp \citep{hu2025finsearchcomprealisticexpertlevelevaluation}, and noisy or conflicting search results in SealQA \citep{pham2026sealqaraisingbarreasoning}. These efforts have substantially advanced both systems and evaluations, but remain largely monolingual or domain-specific, leaving cross-lingual deep research underexplored.

\paragraph{Multilingual and Cross-lingual Retrieval.} Multilingual and cross-lingual retrieval has moved from translation-mediated CLIR toward shared embedding spaces. \textsc{mE5} \citep{wang2024multilinguale5textembeddings} extends the E5 recipe with billion-scale multilingual contrastive pre-training and supervised fine-tuning, while later systems expand the design space through long-context encoders in \textsc{mGTE} \citep{zhang-etal-2024-mgte}, efficiency- and compression-aware multilingual embeddings in \textsc{Arctic-Embed 2.0} \citep{yu2024arcticembed20multilingualretrieval}, and foundation-model-based multilingual training in \textsc{Qwen3 Embedding} \citep{zhang2025qwen3embeddingadvancingtext}. This progress is accompanied by a broader recognition that CLIR is not simply monolingual retrieval plus translation: retrieval quality depends on cross-lingual representation alignment, resource imbalance, domain transfer, and evaluation design \citep{goworek2025bridginglanguagegapsadvances}. Evaluation has therefore expanded to representative benchmarks such as MMTEB \citep{enevoldsen2025mmtebmassivemultilingualtext}, MIRACL \cite{zhang-etal-2023-miracl}, and MLDR \cite{chen-etal-2024-m3}, but it remains a fixed-collection ranking problem. Large-scale CLIR experiments show that multilingual bi-encoders and translation-based lexical retrieval dominate across different datasets and language regimes \citep{zuo-etal-2025-evaluating}; task-specific fact-checking studies further show that multilingual and cross-lingual retrieval yield different model rankings and gains from supervised adaptation \citep{ramponi-etal-2025-multilingual}. These works provide strong retrievers and ranking-oriented evaluations, but not a view of cross-lingual retrieval inside the iterative search, evidence selection, and answer synthesis loop of deep research agents.

\section{Building \textsc{XBCP}}
\subsection{Translation-Based Construction}
We build \textsc{XBCP} by translating the evidence side of BrowseComp-Plus \citep{chen2025browsecompplusfairtransparentevaluation}: questions remain in English, final answers are evaluated in English, and only the evidence documents vary in languages. We use \textsc{GPT-5.4} \citep{openai2026gpt54thinking} as the translation model with a single language-conditioned prompt that requests complete translation into the target language, including titles, terminology, proper nouns, and metadata field names, while preserving URLs, email addresses, formulas, and code blocks; the full prompt is shown in Appendix~\ref{app:translation-prompt}. This prompt is applied to each source document for the non-English target languages, while English documents are retained unchanged. The resulting evidence languages are designed to span different resource conditions. We include relatively high-resource languages with substantial web and retrieval coverage, namely Chinese, English, French, German, Japanese, Korean, Portuguese, and Spanish, as well as low-resource African languages, namely Swahili, Wolof, Yoruba, and Zulu. This language set allows \textsc{XBCP} to test whether cross-lingual deep research systems degrade smoothly across resource regimes or fail disproportionately when evidence appears in languages with weaker retrieval and modeling support.

The translated corpus supports two evaluation configurations. In the cross-lingual setting, each query is assigned to one evidence language, so all supporting documents for that query appear in the same language(English serves as an untranslated reference).Appendix Table~\ref{tab:xbcp-crosslingual-assignments} reports the resulting 830 query assignments and 5,040 evidence-document assignments. In the multilingual setting, 5,040 evidence document instances are randomly but equally assigned to 12 languages, making 420 evidence docs per language; Appendix Table~\ref{tab:xbcp-multilingual-corpora} gives the per-language document counts. This construction lets us vary the linguistic form of the evidence while preserving the original task semantics, making retrieval failures and agent-side synthesis failures comparable across languages.

\subsection{Verification and Quality Control}

To assess the quality of the translated evidence, we conduct an independent expert verification study following the translation-evaluation rubric of MMLU-ProX~\citep{xuan2025mmluproxmultilingualbenchmarkadvanced}. The rubrics is in Appendix~\ref{app:verification-rubrics}.  We sample 200 translated documents from each of 11 non-English languages, yielding 2200 translation instances in total. Expert annotators compare each translation against the original English document and rate it along the same three dimensions in MMLU-ProX, accuracy, fluency, and completeness on 1-5 scale, so that the verification focuses on whether the translated documents preserve the evidence needed for retrieval and answer synthesis. Verification results are in Appendix \ref{app:translation-verification-results}. All language-level mean scores exceed 4.0, suggesting that the translated evidence is generally usable for controlled evaluation, while residual artifacts may remain.

\section{Experiments and Results}
\subsection{Experimental Setup}
Following the evaluation protocol of BrowseComp-Plus \citep{chen2025browsecompplusfairtransparentevaluation}, we evaluate \textsc{XBCP} by pairing search agents with controlled retriever tools over fixed corpora. We consider four agents: \textsc{GPT-OSS-20B} \citep{openai2025gptoss120bgptoss20bmodel}, \textsc{GPT-OSS-120B} \citep{openai2025gptoss120bgptoss20bmodel}, \textsc{Qwen3.6-35B-A3B} \citep{qwen36_35b_a3b}, and \textsc{DeepSeek-V4-Pro} \citep{deepseekai2026deepseekv4}. For retrieval, we compare a sparse lexical baseline, BM25 \citep{BM25}, with four dense multilingual retrievers: \textsc{Qwen3-Embedding-4B}, \textsc{Qwen3-Embedding-8B} \citep{zhang2025qwen3embeddingadvancingtext}, \textsc{Multilingual-E5-Large} \citep{wang2024multilinguale5textembeddings}, and \textsc{Arctic-Embed-L-2.0} \citep{yu2024arcticembed20multilingualretrieval}. \textsc{GPT-OSS-20B}, \textsc{GPT-OSS-120B}, and \textsc{Qwen3.6-35B-A3B} are evaluated with all five retrievers, while \textsc{DeepSeek-V4-Pro} is evaluated with BM25 and \textsc{Qwen3-Embedding-8B}. Each available agent-retriever pair is evaluated on three corpus conditions.

Evaluations are at two complementary levels. First, end-to-end agent performance captures whether an agent can answer correctly while using a retriever as its search tool. Accuracy scores final answer correctness; evidence recall, computed over the union of documents returned across the agent's search trajectory, measures retriever-side coverage of human-verified evidence independent of downstream agent behavior; average search calls captures exploration cost; and calibration error measures the mismatch between the agent's stated confidence and its observed correctness.

Second, we analyze retriever behavior as it appears inside the agent loop. In this setting, retrieval quality is not only a top-$k$ ranking property: a useful retriever should surface supporting documents consistently enough for the agent to find them through iterative search, reduce unnecessary follow-up searches, and provide evidence that can be cited in the final response. We therefore report citation coverage, average citation count, citation precision, and citation recall to measure whether retrieved evidence is carried through into faithful source attribution.

Beyond these two levels, we additionally evaluate an oracle retrieval setting that bypasses search and ranking by supplying all supporting evidence directly to the agent, isolating reasoning errors from retrieval errors. We also report three supplementary analyses: a per-language decomposition, a reasoning-based query expansion experiment, and a reasoning-effort control study.

Since our benchmarks are set in multilingual and crosslingual settings, the original selected models \textsc{Qwen3-32B} \citep{yang2025qwen3technicalreport} and \textsc{GPT-4.1} \citep{openai2025gpt41} in LLM-as-Judge in BrowseComp-Plus are not suitable in our experiments. We therefore adopt \textsc{GPT-5.4} \citep{openai2026gpt54thinking} and change the judge prompt for evaluation. The new judge prompt is in Appendix \ref{app:judge-prompt}.

\subsection{Main Results}
\subsubsection{End-to-End Agent Evaluation}
Table~\ref{tab:e2e-main} reports end-to-end accuracy and evidence recall. The strongest overall performance is obtained by \textsc{DeepSeek-V4-Pro} with \textsc{Qwen3-Embedding-8B}, reaching 64.70\% accuracy on the original corpus, 48.80\% in the multilingual setting, and 42.29\% in the cross-lingual setting. Among the agents evaluated with the full retriever suite, \textsc{Qwen3-Embedding-8B} also gives the strongest original-corpus performance, consistent with the BrowseComp-Plus finding that stronger retrievers improve deep-research agents by surfacing more useful evidence during iterative search \citep{chen2025browsecompplusfairtransparentevaluation}.

The same table shows that translated evidence introduces a large additional difficulty. With \textsc{Qwen3-Embedding-8B}, accuracy drops by roughly 16--23 pp across agents when moving from the original corpus to the translated settings. The degradation appears not only with BM25 but also with dense multilingual retrievers. Meanwhile, multilingual and cross-lingual results are close across most agent--retriever pairs, suggesting that the primary bottleneck is language mismatch rather than the specific language-assignment regime.

\begin{table*}[t]
\centering
\small
\setlength{\tabcolsep}{2.4pt}
\resizebox{\textwidth}{!}{
\begin{tabular}{@{}llrrrrrrrrrr@{}}
\toprule
\multirow{2}{*}{Agent} & \multirow{2}{*}{Retriever}
& \multicolumn{5}{c}{Accuracy (\%)} & \multicolumn{5}{c}{Evidence Recall (\%)} \\
\cmidrule(lr){3-7}\cmidrule(lr){8-12}
& & Orig. & Multi. & $\Delta_M$ & Cross. & $\Delta_C$ & Orig. &  Multi. & $\Delta_M$ & Cross. & $\Delta_C$ \\
\midrule
\multirow{5}{*}{\textsc{GPT-OSS-20B}}
 & BM25 & 15.18 & 3.13 & -12.05 & 3.49 & -11.69 & 22.58 & 5.10 & -17.48 & 5.59 & -16.99 \\
 & \textsc{Qwen3-Embedding-4B} & 29.04 & 11.57 & -17.47 & 11.81 & -17.23 & 38.13 & 23.74 & -14.39 & 23.12 & -15.01 \\
 & \textsc{Qwen3-Embedding-8B} & \textbf{32.89} & 12.05 & -20.84 & \textbf{11.93} & -20.96 & \textbf{42.91} & \textbf{24.60} & -18.31 & \textbf{23.95} & -18.96 \\
 & \textsc{Multilingual-E5-Large} & 20.84 & 4.10 & -16.74 & 3.37 & -17.47 & 24.28 & 4.76 & -19.52 & 4.40 & -19.88 \\
 & \textsc{Arctic-Embed-L-2.0} & 28.80 & \textbf{12.17} & -16.63 & 10.96 & -17.84 & 37.27 & 20.92 & -16.35 & 20.42 & -16.85 \\
\midrule
\multirow{5}{*}{\textsc{GPT-OSS-120B}}
 & BM25 & 22.65 & 5.90 & -16.75 & 5.42 & -17.23 & 31.21 & 9.14 & -22.07 & 8.24 & -22.97 \\
  & \textsc{Qwen3-Embedding-4B} & 35.42 & 13.25 & -22.17 & \textbf{15.30} & -20.12 & 45.55 & 28.53 & -17.02 & 28.50 & -17.05 \\
 & \textsc{Qwen3-Embedding-8B} & \textbf{38.07} & 14.58 & -23.49 & 15.18 & -22.89 & \textbf{48.19} & \textbf{29.85} & -18.34 & \textbf{28.85} & -19.34 \\
 & \textsc{Multilingual-E5-Large} & 20.84 & 6.51 & -14.33 & 5.54 & -15.30 & 25.26 & 6.32 & -18.94 & 6.60 & -18.66 \\
 & \textsc{Arctic-Embed-L-2.0} & 33.61 & \textbf{14.82} & -18.79 & 14.46 & -19.15 & 43.73 & 26.83 & -16.90 & 24.41 & -19.32 \\
\midrule
\multirow{5}{*}{\textsc{Qwen3.6-35B-A3B}}
 & BM25 & 21.69 & 5.42 & -16.27 & 5.18 & -16.51 & 24.12 & 6.91 & -17.21 & 6.10 & -18.02 \\
 & \textsc{Qwen3-Embedding-4B} & 32.29 & \textbf{18.31} & -13.98 & 16.51 & -15.78 & 37.33 & 25.66 & -11.67 & 24.44 & -12.89 \\
 & \textsc{Qwen3-Embedding-8B} & \textbf{38.55} & 18.19 & -20.36 & \textbf{17.95} & -20.60 & \textbf{43.14} & \textbf{28.08} & -15.06 & \textbf{26.74} & -16.40 \\
 & \textsc{Multilingual-E5-Large} & 24.82 & 5.66 & -19.16 & 5.42 & -19.40 & 24.31 & 4.79 & -19.52 & 4.57 & -19.74 \\
 & \textsc{Arctic-Embed-L-2.0} & 35.18 & 17.95 & -17.23 & 15.90 & -19.28 & 37.46 & 21.74 & -15.72 & 20.51 & -16.95 \\
\midrule
\multirow{2}{*}{\textsc{DeepSeek-V4-Pro}}
 & BM25 & 45.06 & 18.55 & -26.51 & 17.47 & -27.59 & 51.08 & 18.15 & -32.93 & 18.51 & -32.57 \\
 & \textsc{Qwen3-Embedding-8B} & \textbf{64.70} & \textbf{48.80} & -15.90 & \textbf{42.29} & -22.41 & \textbf{72.77} & \textbf{59.78} & -12.99 & \textbf{53.82} & -18.95 \\
\bottomrule
\end{tabular}
}
\caption{End-to-end agent performance across corpus conditions.  Multi. denotes the multilingual corpus, Cross. denotes the cross-lingual corpus, and $\Delta_M$ and $\Delta_C$ denote changes from the original corpus to the  multilingual and cross-lingual corpora, respectively. \textsc{DeepSeek-V4-Pro} is evaluated with BM25 and \textsc{Qwen3-Embedding-8B}.}
\label{tab:e2e-main}
\end{table*}

The efficiency and calibration trends reinforce this conclusion. Table~\ref{tab:search-calibration-main} shows that agents generally issue more searches after evidence is translated, but these additional searches do not recover the lost accuracy. Calibration error also increases in both translated settings, indicating that cross-lingual evidence makes agents not only less accurate, but also less reliable in estimating their own correctness.

\begin{table}[t]
\centering
\small
\setlength{\tabcolsep}{5pt}
\begin{tabular}{llrr}
\toprule
Agent & Corpus & Search & Cal.Err. \\
\midrule
\multirow{3}{*}{\textsc{GPT-OSS-20B}} & Original & 13.24 & 34.64 \\
 & Multilingual & 14.20 & 42.25 \\
 & Cross-lingual & 14.16 & 42.56 \\
\midrule
\multirow{3}{*}{\textsc{GPT-OSS-120B}} & Original & 25.35 & 42.50 \\
 & Multilingual & 30.01 & 57.78 \\
 & Cross-lingual & 30.45 & 57.45 \\
\midrule
\multirow{3}{*}{\textsc{Qwen3.6-35B-A3B}} & Original & 13.93 & 36.18 \\
 & Multilingual & 16.07 & 44.06 \\
 & Cross-lingual & 16.31 & 45.90 \\
\midrule
\multirow{3}{*}{\textsc{DeepSeek-V4-Pro}} & Original & 28.52 & 11.94 \\
 & Multilingual & 33.10 & 16.95 \\
 & Cross-lingual & 34.93 & 19.42 \\
\bottomrule
\end{tabular}
\caption{Search efficiency and calibration error with \textsc{Qwen3-Embedding-8B}. Search denotes average search calls per query; calibration error is reported in percentages.}
\label{tab:search-calibration-main}
\end{table}

\subsubsection{Retriever Evaluation}
Evidence recall in Table~\ref{tab:e2e-main} makes the retrieval bottleneck visible. \textsc{Qwen3-Embedding-8B} consistently retrieves the most supporting evidence, while BM25 drops sharply under translated evidence, confirming that lexical matching is poorly suited to English queries over non-English documents. Other dense multilingual retrievers recover part of the loss, but still trail the strongest retriever and remain substantially weaker after translation. Thus, standard multilingual retrieval ability does not directly translate into robust retrieval for complex agentic search.

We further examine whether retrieved evidence is used correctly in final answers. Table~\ref{tab:citation-main} shows that citation coverage, precision, and recall all decline once evidence is translated. This indicates that language mismatch affects not only retrieval, but also whether retrieved sources are carried through into faithful attribution. We provide a citation-error case study in Appendix~\ref{app:citation-case-study}.

\begin{table}[t]
\centering
\small
\setlength{\tabcolsep}{4pt}
\resizebox{\linewidth}{!}{
\begin{tabular}{llrrrr}
\toprule
Agent & Corpus & Cov. & Avg. Cit. & Prec. & Rec. \\
\midrule
\multirow{3}{*}{\textsc{GPT-OSS-20B}} & Original & 50.36 & 2.38 & 66.33 & 29.50 \\
 & Multilingual & 31.93 & 1.91 & 45.99 & 16.65 \\
 & Cross-lingual & 30.36 & 2.01 & 42.58 & 18.78 \\
\midrule
\multirow{3}{*}{\textsc{GPT-OSS-120B}} & Original & 60.60 & 4.11 & 50.89 & 28.99 \\
 & Multilingual & 43.86 & 3.88 & 24.30 & 12.18 \\
 & Cross-lingual & 42.65 & 3.76 & 26.26 & 14.73 \\
\midrule
\multirow{3}{*}{\textsc{Qwen3.6-35B-A3B}} & Original & 41.45 & 3.06 & 72.39 & 40.16 \\
 & Multilingual & 24.82 & 2.78 & 59.55 & 31.80 \\
 & Cross-lingual & 25.66 & 2.74 & 61.11 & 31.34 \\
\midrule
\multirow{3}{*}{\textsc{DeepSeek-V4-Pro}} & Original & 88.07 & 4.03 & 85.80 & 61.30 \\
 & Multilingual & 79.64 & 3.57 & 79.64 & 50.09 \\
 & Cross-lingual & 74.46 & 3.63 & 70.76 & 48.34 \\
\bottomrule
\end{tabular}
}
\caption{Citation behavior with \textsc{Qwen3-Embedding-8B}. Cov., Prec., and Rec. denote citation coverage, citation precision, and citation recall, all in percentages.}
\label{tab:citation-main}
\end{table}

\subsubsection{Oracle Retrieval}
The oracle setting provides a diagnostic decomposition of the end-to-end results. Table~\ref{tab:oracle-main} compares the strongest tool-based condition, using \textsc{Qwen3-Embedding-8B}, with an oracle condition in which all supporting evidence is supplied directly. The retrieval/search gap remains large in every corpus condition: oracle retrieval improves accuracy by over 55 pp on the original corpus and by roughly 65--75 pp after translation. Thus, the largest absolute headroom still lies in getting the right evidence into the agent's context during iterative search.

At the same time, oracle retrieval does not eliminate the cross-lingual penalty. Even with all required evidence provided, translated-evidence oracle accuracy remains below original-corpus oracle accuracy for all agents. These gaps reveal an agent-side bottleneck beyond retrieval: the model must identify relevant facts, align them with the English question, and synthesize an English answer without losing the evidential constraint. We further decompose this bottleneck using a fully target-language oracle variant in Appendix~\ref{app:oracle-tqtp}.

\begin{table}[t]
\centering
\small
\setlength{\tabcolsep}{3.5pt}
\resizebox{\linewidth}{!}{
\begin{tabular}{@{}llrrrr@{}}
\toprule
Agent & Corpus & Tool & Oracle & Ret. Gap & Lang. Gap \\
\midrule
\multirow{3}{*}{\textsc{GPT-OSS-20B}} & Orig. & 32.89 & 90.36 & 57.47 & -- \\
 & Multi. & 12.05 & 81.20 & 69.15 & 9.16 \\
 & Cross. & 11.93 & 77.59 & 65.66 & 12.77 \\
\midrule
\multirow{3}{*}{\textsc{GPT-OSS-120B}} & Orig. & 38.07 & 94.70 & 56.63 & -- \\
 & Multi. & 14.58 & 89.52 & 74.94 & 5.18 \\
 & Cross. & 15.18 & 85.28 & 70.10 & 9.42 \\
\midrule
\multirow{3}{*}{\textsc{Qwen3.6-35B-A3B}} & Orig. & 38.55 & 93.86 & 55.31 & -- \\
 & Multi. & 18.19 & 90.00 & 71.81 & 3.86 \\
 & Cross. & 17.95 & 89.16 & 71.21 & 4.70 \\
\bottomrule
\end{tabular}
}
\caption{Oracle retrieval and error decomposition. Tool accuracy uses \textsc{Qwen3-Embedding-8B}. Ret. Gap is oracle accuracy minus tool-based accuracy under the same corpus condition; Lang. Gap is the drop from original-corpus oracle accuracy to translated-evidence oracle accuracy.}
\label{tab:oracle-main}
\end{table}
\subsection{Supplementary Analyses}
\subsubsection{Per-Language Decomposition}
Table~\ref{tab:perlang-main} reports a per-language decomposition for \textsc{Qwen3.6-35B-A3B} with \textsc{Qwen3-Embedding-8B}, with English as an untranslated reference and the remaining languages grouped by resource level. Full results for other agent–retriever pairs appear in Appendix~\ref{app:per-language-results}.

Two patterns stand out. First, resource level is most visible before oracle retrieval. High-resource languages average 18.39\% tool accuracy and 28.48\% evidence recall, whereas low-resource languages average 10.87\% and 18.00\%, respectively. Yet their oracle accuracies remain relatively close, at 89.67\% and 87.32\%. This suggests that the low-resource penalty in this batch is driven primarily by retrieval failure rather than by an intrinsic inability to answer once evidence is provided. Swahili and Wolof illustrate this most sharply: oracle accuracy stays near 86–90\% while tool-based accuracy collapses to roughly 15\%.

Second, resource level alone does not explain all variations. Within the high-resource group, French, German, Portuguese, and Spanish substantially outperform Japanese and Korean, with Japanese also showing one of the lowest oracle accuracies; Zulu exhibits an analogous pattern among low-resource languages. Cross-lingual deep research is therefore shaped by two separable but interacting factors: the retriever's ability to surface evidence across languages, and the agent's ability to align language-specific evidence with an English query.

\begin{table}[t]
\centering
\small
\setlength{\tabcolsep}{4pt}
\resizebox{\linewidth}{!}{
\begin{tabular}{@{}lrrrrr@{}}
\toprule
Language & $N$ & Tool Acc. & Ev. Rec. & Oracle Acc. & O--T Gap \\
\midrule
\multicolumn{6}{@{}l}{\textit{Untranslated reference}} \\
English & 70 & 42.86 & 49.20 & 92.86 & 50.00 \\
\addlinespace[2pt]
\multicolumn{6}{@{}l}{\textit{High-resource translated languages}} \\
Chinese & 70 & 15.71 & 27.92 & 91.43 & 75.72 \\
French & 69 & 26.09 & 35.15 & 97.10 & 71.01 \\
German & 69 & 27.54 & 37.00 & 94.20 & 66.66 \\
Japanese & 69 & 4.35 & 12.29 & 73.91 & 69.56 \\
Korean & 69 & 10.14 & 17.71 & 85.51 & 75.37 \\
Portuguese & 69 & 23.19 & 29.43 & 95.65 & 72.46 \\
Spanish & 69 & 21.74 & 39.86 & 89.86 & 68.12 \\
\addlinespace[2pt]
\multicolumn{6}{@{}l}{\textit{Low-resource translated languages}} \\
Swahili & 69 & 17.39 & 23.70 & 89.86 & 72.47 \\
Wolof & 69 & 14.49 & 20.50 & 86.96 & 72.47 \\
Yoruba & 69 & 7.25 & 15.39 & 94.20 & 86.95 \\
Zulu & 69 & 4.35 & 12.41 & 78.26 & 73.91 \\
\midrule
\textit{High-resource avg.} & 484 & 18.39 & 28.48 & 89.67 & 71.28 \\
\textit{Low-resource avg.} & 276 & 10.87 & 18.00 & 87.32 & 76.45 \\
\bottomrule
\end{tabular}
}
\caption{Per-language results in the cross-lingual setting for \textsc{Qwen3.6-35B-A3B} with \textsc{Qwen3-Embedding-8B}, plus oracle accuracy for the same agent. All scores are percentages except $N$. O--T Gap denotes oracle accuracy minus tool-based accuracy. Group averages are weighted by the number of queries and exclude the untranslated English reference.}
\label{tab:perlang-main}
\end{table}

\subsubsection{The Impact of Query Expansion}
\citet{chen2026agentirreasoningawareretrievaldeep} argue that deep research agents expose a retrieval signal that conventional retrievers ignore: before issuing a search query, the agent often writes a natural-language reasoning trace that clarifies the task intent, summarizes prior findings, and identifies unresolved evidence needs. Their full \textsc{AgentIR} system trains a retriever to jointly embed the reasoning trace and the issued query. We study a lighter-weight variant in \textsc{XBCP}: without any retriever training or index changes, we use the agent's current reasoning trace as query expansion by concatenating it with the issued search query before passing the input to \textsc{Qwen3-Embedding-8B}. This isolates whether agent-side reasoning is already useful as a retrieval signal, and whether the benefit survives when the relevant evidence is written in another language.

\noindent\begin{minipage}{\columnwidth}
\centering
\small
\setlength{\tabcolsep}{4pt}
\begin{tabular}{llrrrr}
\toprule
Corpus & Method & Acc. & Ev. Rec. & Cal.Err. & Search \\
\midrule
\multirow{3}{*}{Orig.}
 & Standard & 32.89 & 42.91 & 34.64 & 13.24 \\
 & +Reason. & 36.14 & 47.77 & 34.02 & 12.96 \\
 & \textit{$\Delta$} & \textit{+3.25} & \textit{+4.86} & \textit{-0.62} & \textit{-0.28} \\
\midrule
\multirow{3}{*}{Multi.}
 & Standard & 12.05 & 24.60 & 42.25 & 14.20 \\
 & +Reason. & 14.10 & 27.55 & 41.86 & 14.18 \\
 & \textit{$\Delta$} & \textit{+2.05} & \textit{+2.95} & \textit{-0.39} & \textit{-0.02} \\
\midrule
\multirow{3}{*}{Cross.}
 & Standard & 11.93 & 23.95 & 42.56 & 14.16 \\
 & +Reason. & 14.60 & 27.00 & 41.33 & 13.92 \\
 & \textit{$\Delta$} & \textit{+2.67} & \textit{+3.05} & \textit{-1.23} & \textit{-0.24} \\
\bottomrule
\end{tabular}
\captionof{table}{\textsc{AgentIR}-style zero-training query expansion for \textsc{GPT-OSS-20B} with \textsc{Qwen3-Embedding-8B}. +Reason. denotes concatenating the agent's current reasoning trace with the issued query. Acc., Ev. Rec., and Cal.Err. are percentages; Search denotes average search calls per query.}
\label{tab:agentir-expansion}
\end{minipage}

Table~\ref{tab:agentir-expansion} shows that reasoning-based expansion consistently improves performance across all three corpus conditions. On the original corpus, accuracy increases by 3.25 pp and evidence recall by 4.86 pp, while calibration error and search turns both decrease. The same pattern holds after translation, although with smaller gains. The improvements therefore do not come from additional exploration, since the expanded runs use slightly fewer search calls on average; rather, the reasoning trace appears to make each search query more informative.

From the perspective of \textsc{XBCP}, this result has two implications. First, cross-lingual deep research should treat query formulation as part of the retrieval problem: the agent's reasoning can help disambiguate underspecified sub-queries and expose more supporting evidence even without retriever fine-tuning. Second, the smaller gains under translated evidence show that reasoning-aware query expansion is not sufficient by itself. The system still depends on the retriever's cross-lingual alignment to bridge the language gap.

\subsubsection{The Impact of Reasoning Effort}
Following BrowseComp-Plus \citep{chen2025browsecompplusfairtransparentevaluation}, we further examine how reasoning effort affects both answer quality and search behavior. This is a particularly important diagnostic for agentic search: increasing the inference budget may improve final reasoning, but it can also change search iterations and exposed evidence before answering. We therefore vary the reasoning-effort mode of \textsc{GPT-OSS-20B} while holding the retriever fixed to \textsc{Qwen3-Embedding-8B}. This setup asks whether cross-lingual failures can be mitigated by deeper deliberation, or whether language mismatch persists regardless of search effort.

\begin{table}[t]
\centering
\small
\setlength{\tabcolsep}{4pt}
\resizebox{\linewidth}{!}{
\begin{tabular}{llrrrr}
\toprule
Effort & Corpus & Acc. & Ev. Rec. & Search & Cal.Err. \\
\midrule
\multirow{3}{*}{Low}
 & Orig. & 15.18 & 18.16 & 2.05 & 35.67 \\
 & Multi. & 4.58 & 9.56 & 2.01 & 44.58 \\
 & Cross. & 4.94 & 9.33 & 2.01 & 43.94 \\
\midrule
\multirow{3}{*}{Medium}
 & Orig. & 32.89 & 42.91 & 13.24 & 34.64 \\
 & Multi. & 12.05 & 24.60 & 14.20 & 42.25 \\
 & Cross. & 11.93 & 23.95 & 14.16 & 42.56 \\
\midrule
\multirow{3}{*}{High}
 & Orig. & 36.02 & 52.53 & 26.31 & 23.36 \\
 & Multi. & 15.30 & 34.09 & 29.31 & 36.02 \\
 & Cross. & 15.18 & 33.44 & 28.66 & 37.42 \\
\bottomrule
\end{tabular}
}
\caption{Impact of reasoning effort for \textsc{GPT-OSS-20B} with \textsc{Qwen3-Embedding-8B}. Acc., Ev. Rec., and Cal.Err. are percentages; Search denotes average search calls per query.}
\label{tab:reasoning-effort}
\end{table}

Table~\ref{tab:reasoning-effort} shows that higher reasoning effort consistently improves both accuracy and evidence recall. From low to high effort, an increase from 15.18\% to 36.02\% is observed for the original corpus, and over 10 pp increases are observed for both translated settings. Evidence recall follows the same pattern, increasing in all 3 settings. These gains come with a clear efficiency cost: high effort requires over 26 search calls per query, compared with roughly 2 calls under low effort. Calibration also improves at high effort, suggesting that more extensive search and deliberation make the agent less overconfident.

The comparison with the original corpus is more revealing. High-effort cross-lingual and multilingual runs reach only about the accuracy of the low-effort original run, despite using more than 14 times as many search calls; they remain far below the medium-effort original run. Thus, additional reasoning effort improves the agent in every corpus condition, but it does not turn cross-lingual evidence into a monolingual problem. In conclusion, the dominant difficulty is the language mismatch between the English information need and translated evidence, rather than the specific corpus assignment regime.

\section{Discussion}
Our experiments identify cross-linguality as a structural source of difficulty for deep research agents, not merely as a perturbation to first-stage retrieval. By varying only the evidence language, \textsc{XBCP} isolates how language mismatch propagates through the evidence-seeking pipeline. This design brings together two evaluation traditions that have largely remained separate. Multilingual and cross-lingual retrieval benchmarks \citep{zhang-etal-2023-miracl,enevoldsen2025mmtebmassivemultilingualtext,zuo-etal-2025-evaluating} isolate whether a system can rank relevant documents across languages in a fixed collection, while deep research benchmarks \citep{wei2025browsecompsimplechallengingbenchmark,chen2025browsecompplusfairtransparentevaluation} evaluate iterative evidence seeking and grounded answer synthesis but typically assume that questions and evidence are linguistically aligned. \textsc{XBCP} connects these views by asking whether cross-lingual retrieval remains effective once it becomes part of an agentic search process.

This perspective first reveals a retrieval/search bottleneck. Our results show that dense multilingual retrievers outperform BM25 after translation. Yet conventional retrieval success does not guarantee that an agent will find the right evidence during iterative search. This gap is consistent with prior work showing that multilingual and cross-lingual retrieval can exhibit different behavior across language regimes and retrieval configurations \citep{ramponi-etal-2025-multilingual,zuo-etal-2025-evaluating,zeng2026codeswitchinginformationretrievalbenchmarks}. In \textsc{XBCP}, the same issue appears inside the agent loop: translated corpora reduce evidence recall, increase search effort, and lower citation reliability even when the retriever is dense and multilingual. The implication is that cross-lingual retrievers should not be evaluated only by whether they rank relevant documents highly in isolation, but also by whether they expose the evidence at the right point in an agent's search trajectory.

\textsc{XBCP} also separates this retrieval/search bottleneck from an evidence-integration bottleneck. Recent work on multilingual and cross-lingual RAG has shown that language-mismatched evidence can complicate retrieval, consistency, and reasoning over multilingual contexts \citep{liu-etal-2025-xrag,ranaldi-etal-2026-multilingual,qi2026crosearchr1betterleveragingcrosslingual}. However, these studies remain focused on relatively short-chain single-hop or multi-hop QA settings, leaving the long-horizon deep research setting underexamined. Our oracle results instantiate this distinction: providing all gold evidence substantially raises accuracy, confirming that finding evidence is a major bottleneck, but translated oracle accuracy remains below original one. Thus, cross-lingual deep research is decomposable into two linked questions: whether system can find language-mismatched evidence, and whether it can use evidence faithfully once it is found. The latter requires the agent to identify relevant facts in non-English sources, align them with an English question and answer space, and preserve the evidential constraint during synthesis.

The per-language results further suggest that low-resource effects enter the system primarily before evidence reaches the model. Multilingual retrieval evaluation has long emphasized that language resource level, typology, and annotation coverage shape retrieval behavior \citep{zhang-etal-2023-miracl}; multilingual LLM research similarly identifies language imbalance and multilingual alignment as central challenges \citep{Xu_2025}. In \textsc{XBCP}, low-resource languages show substantially lower tool-based accuracy and evidence recall than high-resource languages, but their oracle accuracy is comparable. It indicates that  the largest low-resource penalty appears during retrieval: once strong agents receive the gold documents, they can still extract and integrate the relevant information. Resource-level effects enter the system primarily before evidence reaches the model.

Taken together, these findings point toward language-aware agentic search rather than simply stronger multilingual retrieval. Active retrieval work argues that systems should decide dynamically when and what to retrieve during generation \citep{jiang-etal-2023-active}, while CLIR research has increasingly moved from translation-based methods toward LLM-based alignment, with cross-lingual representation alignment remaining a central challenge \citep{goworek2025bridginglanguagegapsadvances}. \textsc{XBCP} extends this view to deep research: agents need to recognize the language of available evidence, formulate queries across languages and entity variants, decide when translation or language-specific search is needed, and preserve source attribution across the final answer. Cross-lingual deep research therefore requires coordination between the retriever, query planner, reader, and citation mechanism, so that language-mismatched evidence can be found, interpreted, and cited as part of a single grounded reasoning process.

\section*{Limitations}

Our main experiments report a single evaluation run per agent–retriever–corpus configuration. Running agents over full search trajectories with multiple retrievers across three corpus conditions is computationally expensive, and we did not repeat each configuration over multiple random seeds. While the gaps between corpus conditions and between retrievers are large and consistent across agents, formal variance estimates and significance tests over multiple runs are left to future work.

We use a single set of inference hyperparameters per agent, following each model's recommended generation configuration, without tuning sampling temperature or top-p. This keeps comparisons across conditions controlled, but condition-specific tuning, particularly for low-resource languages, may partially reduce the observed gaps. A systematic study of inference configuration is beyond the scope of this work.
\bibliography{custom}

\clearpage
\raggedbottom
\appendix

\section{\textsc{XBCP} Construction Details}
\label{app:xbcp-construction}
\captionsetup{hypcap=false}

\noindent\begin{minipage}{\columnwidth}
\centering
\small
\begin{tabular}{lrr}
\toprule
Language & Queries & Evidence Docs \\
\midrule
Chinese & 70 & 416 \\
English & 70 & 421 \\
French & 69 & 401 \\
German & 69 & 431 \\
Japanese & 69 & 411 \\
Korean & 69 & 419 \\
Portuguese & 69 & 443 \\
Spanish & 69 & 397 \\
Swahili & 69 & 428 \\
Wolof & 69 & 412 \\
Yoruba & 69 & 443 \\
Zulu & 69 & 418 \\
\midrule
\textbf{Total} & \textbf{830} & \textbf{5,040} \\
\bottomrule
\end{tabular}
\captionof{table}{Language assignment statistics for the cross-lingual setting.}
\label{tab:xbcp-crosslingual-assignments}
\end{minipage}

\medskip
\noindent\begin{minipage}{\columnwidth}
\centering
\small
\begin{tabular}{lr}
\toprule
Language & Documents \\
\midrule
Chinese & 420 \\
English & 420 \\
French & 420 \\
German & 420 \\
Japanese & 420 \\
Korean & 420 \\
Portuguese & 420 \\
Spanish & 420 \\
Swahili & 420 \\
Wolof & 420 \\
Yoruba & 420 \\
Zulu & 420 \\
\midrule
\textbf{Total} & \textbf{5,040} \\
\bottomrule
\end{tabular}
\captionof{table}{Per-language corpus coverage in the multilingual setting. The 420 English source documents are retained unchanged, and the remaining 4,620 document instances are produced by translation.}
\label{tab:xbcp-multilingual-corpora}
\end{minipage}

\section{Translation Prompt}
\label{app:translation-prompt}

\begin{tcolorbox}[
  colback=gray!4,
  colframe=black!35,
  boxrule=0.4pt,
  arc=1mm,
  left=4pt,
  right=4pt,
  top=4pt,
  bottom=4pt,
  title={Prompt used for document translation}
]
\small
\textbf{Instruction.} Translate the following document completely into \texttt{\{target\_language\}}.

Translate everything including proper nouns, titles, terminology, and metadata field names according to \texttt{\{target\_language\}} conventions. For example, \texttt{name:} should become the equivalent in \texttt{\{target\_language\}}, \texttt{birth\_date:} should become the equivalent in \texttt{\{target\_language\}}, etc.

\textbf{Rules.}
\begin{enumerate}
    \item Ensure cultural appropriateness for \texttt{\{target\_language\}} speakers.
    \item If works such as books, movies, TV shows, songs, or other literary/entertainment titles have well-known translations in \texttt{\{target\_language\}}, use those established translations.
    \item Preserve all URLs, email addresses, math formulas, and code blocks unchanged.
    \item Output only the translated document. Do not add explanations.
\end{enumerate}
\end{tcolorbox}

\section{Translation Verification Rubrics}

This translation verification rubrics follows the rubrics conducted by MMLU-ProX \citep{xuan2025mmluproxmultilingualbenchmarkadvanced}.
\label{app:verification-rubrics}

\begin{tcolorbox}[
  breakable,
  colback=gray!4,
  colframe=black!35,
  boxrule=0.4pt,
  arc=1mm,
  left=4pt,
  right=4pt,
  top=4pt,
  bottom=4pt,
  title={Prompt used for expert translation verification}
]
\small
\textbf{Instruction.} You are an expert bilingual evaluator. Compare the source document with its machine-translated version in the target language. Rate the translation on accuracy, fluency, and completeness using the criteria below. Provide a score from 1 to 5 for each dimension and a brief justification for any score below 5.

\textbf{Evaluation Criteria for Expert Rating of Machine Translation Results}\\
\textbf{1. Accuracy (1-5):}

\begin{itemize}
    \item \textbf{5 (Highly Accurate):}
    \begin{itemize}
        \item All key terms and concepts are translated correctly with no errors.
        \item Every technical term corresponds precisely to the original text, with no mistranslations or incorrect word choices.
        \item The most appropriate and professional terminology in the target language is used.
        \item Expressions align with commonly used terminology in professional or technical contexts.
    \end{itemize}

    \item \textbf{4 (Accurate):}
    \begin{itemize}
        \item Most terms and concepts are translated correctly, with only a few minor errors that do not affect overall comprehension.
        \item Some terms may be slightly imprecise, but the translation remains generally accurate.
        \item Uses appropriate terminology in the target language in most cases.
        \item A few terms may be simplified but remain understandable within the intended domain.
    \end{itemize}

    \item \textbf{3 (Moderately Accurate):}
    \begin{itemize}
        \item Key terms and concepts are mostly correct but contain some errors that may cause partial misunderstandings.
        \item Some critical terms are inaccurately translated, requiring the reader to infer the intended meaning.
        \item Slight deviations in the use of target-language terminology.
        \item Occasionally uses uncommon or outdated terms.
    \end{itemize}

    \item \textbf{2 (Somewhat Inaccurate):}
    \begin{itemize}
        \item Many key terms and concepts are mistranslated, significantly affecting comprehension.
        \item Important concepts are incorrectly translated, leading to potential misunderstandings of the original text.
        \item Uses incorrect or inappropriate terminology in the target language.
        \item Terminology is inconsistent, reducing the text’s professionalism.
    \end{itemize}

    \item \textbf{1 (Inaccurate):}
    \begin{itemize}
        \item Frequent and severe mistranslations of key terms and concepts, failing to convey the original meaning.
        \item Most of the content does not match the original text.
        \item Lacks proper use of target-language terminology.
        \item Terminology is chaotic, possibly using irrelevant or incorrect vocabulary entirely.
    \end{itemize}
\end{itemize}

\textbf{2. Fluency (1–5):}

\begin{itemize}
    \item \textbf{5 (Highly Fluent):}
    \begin{itemize}
        \item The target-language expression is natural and smooth, making it effortless to read.
        \item The language style is refined and appropriate for professional or formal contexts.
        \item The sentence structure fully adheres to natural conventions in the target language, with no grammatical or lexical errors.
    \end{itemize}

    \item \textbf{4 (Fluent):}
    \begin{itemize}
        \item The target-language expression is generally natural, with only minor linguistic imperfections that do not affect comprehension.
        \item Some sentences may sound slightly stiff.
        \item Sentence structures mostly conform to target-language norms, with very few grammatical errors.
    \end{itemize}

    \item \textbf{3 (Moderately Fluent):}
    \begin{itemize}
        \item The target-language expression is somewhat unnatural, requiring the reader to adjust their understanding slightly.
        \item Some inappropriate word choices or rigid sentence structures are present.
        \item Sentence structures are mostly correct, but some grammatical errors exist.
    \end{itemize}

    \item \textbf{2 (Somewhat Unnatural):}
    \begin{itemize}
        \item The target-language expression lacks fluency, making it difficult to read smoothly.
        \item Sentence transitions are awkward, and logical connections are unclear.
        \item Many structural issues exist, with frequent grammatical errors.
    \end{itemize}

    \item \textbf{1 (Not Fluent):}
    \begin{itemize}
        \item The target-language expression is highly unnatural or difficult to understand.
        \item Literal translation is evident, lacking natural phrasing in the target language.
        \item The sentence structure is disorganized, with severe grammatical mistakes, making the text unreadable.
    \end{itemize}
\end{itemize}

\textbf{3. Completeness (1–5):}

\begin{itemize}
    \item \textbf{5 (Fully Complete):}
    \begin{itemize}
        \item The full meaning of the original text is retained with no omissions or additions.
        \item All details, data, and annotations are accurately conveyed.
        \item The translation maintains the same length and depth as the original text.
    \end{itemize}

    \item \textbf{4 (Complete):}
    \begin{itemize}
        \item The primary meaning of the original text is retained, with only a few minor details omitted or slightly unclear.
        \item Some less critical information may be left out.
        \item The translation generally corresponds to the original content.
    \end{itemize}

    \item \textbf{3 (Moderately Complete):}
    \begin{itemize}
        \item Most of the original meaning is conveyed, but some information is missing or added.
        \item Important details may be overlooked.
        \item The translation differs from the original in certain aspects, requiring readers to infer some content.
    \end{itemize}

    \item \textbf{2 (Somewhat Incomplete):}
    \begin{itemize}
        \item The core information from the original text is not fully conveyed, with noticeable omissions or unnecessary additions.
        \item Potential inclusion of unrelated information.
        \item The translation does not fully correspond to the original, affecting comprehension.
    \end{itemize}

    \item \textbf{1 (Incomplete):}
    \begin{itemize}
        \item Significant omissions or added incorrect information prevent an accurate reflection of the original text.
        \item Important sections or sentences are missing.
        \item The translation deviates heavily from the original, making it difficult to understand the intended meaning.
    \end{itemize}
\end{itemize}

\end{tcolorbox}

\clearpage

\section{Translation Verification Results}
\label{app:translation-verification-results}

 Table \ref{tab:per-language-translation-verification-results} in this appendix report per-language translation results in our corpora. For each language in translation, we adopt three dimensions in evaluation: accuracy, fluency and completeness. Each language evaluation has 200 samples and the results are reported in average value.

\begin{table}[!htbp]
\centering
\small
\setlength{\tabcolsep}{2.8pt}
\begin{tabular}{@{}lrccc@{}}
\toprule
Language & Accuracy & Fluency & Completeness  \\
\midrule
Chinese & 4.82 & 4.88 & 4.74  \\
French & 4.91 & 4.90 & 4.86 \\
German & 4.74 & 4.78 & 4.68 \\
Japanese & 4.44 & 4.66 & 4.48 \\
Korean & 4.46 & 4.48 & 4.52 \\
Portuguese & 4.89 & 4.88 & 4.88 \\
Spanish & 4.87 & 4.86 & 4.84 \\
Swahili & 4.52 & 4.68 & 4.70 \\
Wolof & 4.18 & 4.34 & 4.38 \\
Yoruba & 4.12 & 4.42 & 4.36 \\
Zulu & 4.10 & 4.22 & 4.28 \\
\bottomrule
\end{tabular}

\caption{Per-language translation verification results. All values are on 1–5 scale.}
\label{tab:per-language-translation-verification-results}
\end{table}

\section{Judge Prompt}
\label{app:judge-prompt}

\begin{tcolorbox}[
  colback=gray!4,
  colframe=black!35,
  boxrule=0.4pt,
  arc=1mm,
  left=4pt,
  right=4pt,
  top=4pt,
  bottom=4pt,
  title={Prompt used for LLM-as-Judge}
]
\small
Judge whether the following \texttt{[response]} to \texttt{[question]} is correct or not based on the precise and unambiguous \texttt{[correct\_answer]} below.

\smallskip
\texttt{[question]}: \texttt{\{question\}}

\texttt{[response]}: \texttt{\{response\}}

\texttt{[correct\_answer]}: \texttt{\{correct\_answer\}}

\smallskip
The evidence documents used to answer this question are in another language. As a result, the extracted answer may be written in another language rather than English. The \texttt{[correct\_answer]} is in English. You must judge whether the extracted answer and the correct answer refer to the same entity, concept, or value, regardless of language differences. For example, ``\ja{ハーバード大学}'' and Harvard University, ``\zh{迈克尔·乔丹}`` and ``Michael Jordan'', or ``\ko{서울}'' and ''Seoul'' should be considered equivalent.

\smallskip
Your judgement must be in the format and criteria specified below:
\begin{enumerate}
    \item \texttt{extracted\_final\_answer}: The final exact answer extracted from the \texttt{[response]}. Put the extracted answer as ``None'' if there is no exact, final answer to extract from the response.
    \item \texttt{reasoning}: Explain why the \texttt{extracted\_final\_answer} is correct or incorrect based on \texttt{[correct\_answer]}, focusing only on whether they refer to the same entity or value. If they are in different languages, determine whether they are translations or transliterations of each other. Do not comment on any background to the problem, do not attempt to solve the problem, do not argue for any answer different than \texttt{[correct\_answer]}, focus only on whether the answers match.
    \item \texttt{correct}: Answer ``yes'' if \texttt{extracted\_final\_answer} matches the \texttt{[correct\_answer]} given above, or is a translation/transliteration of it, or is within a small margin of error for numerical problems. Answer ``no'' otherwise, i.e. if there is any inconsistency, ambiguity, non-equivalency, or if the extracted answer is incorrect.
    \item \texttt{confidence}: The extracted confidence score between 0\% and 100\% from \texttt{[response]}. Put 100 if there is no confidence score available.
\end{enumerate}
\end{tcolorbox}

\section{Decomposing the Agent Cross-lingual Bottleneck}
\label{app:oracle-tqtp}

Our oracle experiments show that providing gold evidence directly to the agent does not fully recover monolingual performance, revealing an \emph{agent-side} cross-lingual bottleneck. A natural follow-up question is whether this bottleneck arises because the agent must reason over non-English evidence, or because it must also switch between an English prompt and non-English content. To disentangle these factors, we introduce a fully target-language oracle variant (\textsc{Oracle-tq+tp}), in which the system prompt, the query, and the evidence documents are all presented in the target language. This removes any language switching and tests whether a monolingual non-English environment helps the agent reason more effectively.

\begin{table}[h]
\centering
\small
\setlength{\tabcolsep}{4pt}
\begin{tabular}{lcc}
\toprule
Oracle Variant & GPT-OSS-20B & GPT-OSS-120B \\
\midrule
EN Oracle (en$\to$en) & 90.36 & 94.70 \\
Oracle (en$\to$xx) & 77.59 & 85.28 \\
Oracle-tq+tp (xx$\to$xx) & 71.67 & 79.90 \\
\bottomrule
\end{tabular}
\caption{Oracle accuracy (\%) on the cross-lingual corpus under three prompt--evidence language configurations. EN Oracle: English prompt + English evidence (upper bound). Oracle: English prompt + target-language evidence. Oracle-tq+tp: target-language prompt + target-language evidence.}
\label{tab:oracle-tqtp}
\end{table}

Table~\ref{tab:oracle-tqtp} shows that, contrary to our expectation, Oracle-tq+tp performs \emph{worse} than the standard Oracle with English prompt: GPT-OSS-20B drops by 5.92\,pp and GPT-OSS-120B by 5.38\,pp. The agent reasons less effectively when the prompt is also in the target language, even though language switching is eliminated. This reveals that the agent's cross-lingual weakness has two distinct components:

\begin{enumerate}[leftmargin=*,nosep]
\item \textbf{Evidence understanding bottleneck} (EN Oracle $\to$ Oracle): the agent loses 12.77\,pp (20B) / 9.42\,pp (120B) from reading non-English evidence, even under English instructions.
\item \textbf{Prompt language penalty} (Oracle $\to$ Oracle-tq+tp): switching the prompt to the target language costs an additional 5.92\,pp (20B) / 5.38\,pp (120B), indicating that these models follow instructions more reliably in English.
\end{enumerate}

These results have two implications. First, the agent bottleneck is \emph{intrinsic} to the model's multilingual reasoning capability, not a surface-level language-switching artifact. Providing a fully monolingual target-language environment does not help; it makes things worse. Second, English serves as the agent's ``native language'' for instruction following: even when all content is non-English, the agent benefits from receiving its task description in English. This suggests that improving cross-lingual agent performance requires stronger multilingual pretraining, not prompt translation.

\section{Citation Precision Error Analysis}
\label{app:citation-case-study}

\textsc{GPT-OSS-120B} exhibits the steepest citation precision drop among all agents: from 50.89\% on the original corpus to 24.30\% (multilingual) and 26.26\% (cross-lingual), a reduction of roughly 50\% (Table~\ref{tab:citation-main}). To diagnose this degradation, we classify every query where \textsc{GPT-OSS-120B} made citations but failed to cite any gold evidence document into two mutually exclusive error types: (1)~the agent retrieved at least one gold document but cited other documents instead (\emph{mapping failure}); (2)~no gold document was retrieved and the agent cited English negative documents instead (\emph{no gold retrieved}). We include \textsc{GPT-OSS-20B} and \textsc{Qwen3.6-35B-A3B} as reference points in Table~\ref{tab:citation-error-all}.

\begin{table}[h]
\centering
\small
\setlength{\tabcolsep}{3pt}
\resizebox{\columnwidth}{!}{
\begin{tabular}{@{}llrrrr@{}}
\toprule
Agent & Corpus & Prec. & Errors & Map.Fail & No Gold \\
\midrule
\multirow{3}{*}{\textsc{GPT-OSS-120B}}
 & Orig. & 50.89 & 226 & 97 (42.92\%) & 129 (57.08\%) \\
 & Multi. & 24.30 & 272 & 104 (38.24\%) & 168 (61.76\%) \\
 & Cross. & 26.26 & 248 & 84 (33.87\%) & 164 (66.13\%) \\
\midrule
\multirow{3}{*}{\textsc{GPT-OSS-20B}}
 & Orig. & 66.33 & 172 & 85 (49.42\%) & 87 (50.58\%) \\
  & Multi. & 45.99 & 165 & 57 (34.55\%) & 108 (65.45\%) \\
  & Cross. & 42.58 & 166 & 72 (43.37\%) & 94 (56.63\%) \\
\midrule
\multirow{3}{*}{\textsc{Qwen3.6-35B}}
 & Orig. & 72.39 & 108 & 50 (46.30\%) & 58 (53.70\%) \\
 & Multi. & 59.55 & 98 & 29 (29.59\%) & 69 (70.41\%) \\
 & Cross. & 61.11 & 90 & 22 (24.44\%) & 68 (75.56\%) \\

\bottomrule
\end{tabular}
}
\caption{Citation error classification with \textsc{Qwen3-Embedding-8B}. Prec.\ is citation precision (\%) among queries with citations. Errors is the number of queries that cited zero gold documents. Map.Fail: gold was retrieved but agent cited other documents. No Gold: no gold document was retrieved. Percentages in parentheses sum to 100\% within each row.}
\label{tab:citation-error-all}
\end{table}

For \textsc{GPT-OSS-120B}, the dominant error type is \emph{no gold retrieved}, accounting for 57.08\% of errors on the original corpus and rising to 66.13\% on the multilingual corpus. In these cases, the retriever never surfaced the gold document during the agent's search trajectory, so the agent cited English negative documents that appeared topically related but did not contain the correct evidence. Mapping failures account for the remaining 33.87--42.92\% of errors and decline as a share after translation, not because the agent improves at citation mapping, but because fewer gold documents are retrieved in the first place.

Compared with \textsc{GPT-OSS-20B} and \textsc{Qwen3.6-35B-A3B}, \textsc{GPT-OSS-120B} has substantially more total errors (226--272 vs.\ 108--172). This is driven by its higher citation coverage (60.6\% vs.\ 50.4\% and 41.5\%): the 120B model cites documents more frequently, creating more opportunities for incorrect citations.
\section{Additional Per-Language Results}
\label{app:per-language-results}

All tables in this appendix report per-language results in the cross-lingual setting. Q3-4B and Q3-8B denote \textsc{Qwen3-Embedding-4B} and \textsc{Qwen3-Embedding-8B}, respectively.

\begin{table}[!htbp]
\centering
\scriptsize
\setlength{\tabcolsep}{2.8pt}
\resizebox{\columnwidth}{!}{
\begin{tabular}{@{}lrccccc@{}}
\toprule
Language & $N$ & BM25 & Q3-4B & Q3-8B & E5 & Arctic \\
\midrule
Chinese & 70 & 0.00 & 8.57 & 8.57 & 1.43 & 7.14 \\
English & 70 & 17.14 & 27.14 & 42.86 & 20.00 & 30.00 \\
French & 69 & 4.35 & 13.04 & 15.94 & 0.00 & 17.39 \\
German & 69 & 4.35 & 33.33 & 15.94 & 1.45 & 18.84 \\
Japanese & 69 & 0.00 & 1.45 & 1.45 & 2.90 & 1.45 \\
Korean & 69 & 4.35 & 4.35 & 5.80 & 4.35 & 1.45 \\
Portuguese & 69 & 4.35 & 18.84 & 18.84 & 2.90 & 17.39 \\
Spanish & 69 & 1.45 & 15.94 & 13.04 & 1.45 & 26.09 \\
Swahili & 69 & 2.90 & 10.14 & 10.14 & 2.90 & 1.45 \\
Wolof & 69 & 0.00 & 5.80 & 5.80 & 1.45 & 2.90 \\
Yoruba & 69 & 1.45 & 2.90 & 2.90 & 0.00 & 7.25 \\
Zulu & 69 & 1.45 & 0.00 & 1.45 & 1.45 & 0.00 \\
\midrule
\textbf{Total} & 830 & 3.49 & 11.81 & 11.93 & 3.37 & 10.96 \\
\bottomrule
\end{tabular}
}
\caption{Per-language tool-based accuracy for \textsc{GPT-OSS-20B}. All values are percentages.}
\label{tab:perlang-gptoss-tool}
\end{table}

\begin{table}[!htbp]
\centering
\scriptsize
\setlength{\tabcolsep}{2.8pt}
\resizebox{\columnwidth}{!}{
\begin{tabular}{@{}lrccccc@{}}
\toprule
Language & $N$ & BM25 & Q3-4B & Q3-8B & E5 & Arctic \\
\midrule
Chinese & 70 & 1.43 & 10.00 & 5.71 & 1.43 & 10.00 \\
English & 70 & 30.00 & 38.57 & 40.00 & 34.29 & 37.14 \\
French & 69 & 7.25 & 24.64 & 18.84 & 2.90 & 17.39 \\
German & 69 & 5.80 & 33.33 & 26.09 & 1.45 & 24.64 \\
Japanese & 69 & 2.90 & 1.45 & 2.90 & 1.45 & 5.80 \\
Korean & 69 & 4.35 & 2.90 & 8.70 & 2.90 & 7.25 \\
Portuguese & 69 & 2.90 & 26.09 & 20.29 & 5.80 & 26.09 \\
Spanish & 69 & 2.90 & 14.49 & 21.74 & 5.80 & 23.19 \\
Swahili & 69 & 2.90 & 13.04 & 17.39 & 4.35 & 11.59 \\
Wolof & 69 & 2.90 & 7.25 & 5.80 & 4.35 & 1.45 \\
Yoruba & 69 & 1.45 & 5.80 & 7.25 & 0.00 & 7.25 \\
Zulu & 69 & 0.00 & 5.80 & 7.25 & 1.45 & 1.45 \\
\midrule
\textbf{Total} & 830 & 5.42 & 15.30 & 15.18 & 5.54 & 14.46 \\
\bottomrule
\end{tabular}
}
\caption{Per-language tool-based accuracy for \textsc{GPT-OSS-120B}. All values are percentages.}
\label{tab:perlang-gptoss120b-tool}
\end{table}

\begin{table}[!htbp]
\centering
\scriptsize
\setlength{\tabcolsep}{2.8pt}
\resizebox{\columnwidth}{!}{
\begin{tabular}{@{}lrccccc@{}}
\toprule
Language & $N$ & BM25 & Q3-4B & Q3-8B & E5 & Arctic \\
\midrule
Chinese & 70 & 1.43 & 15.71 & 15.71 & 2.86 & 17.14 \\
English & 70 & 18.57 & 42.86 & 42.86 & 25.71 & 35.71 \\
French & 69 & 5.80 & 18.84 & 26.09 & 1.45 & 18.84 \\
German & 69 & 10.14 & 31.88 & 27.54 & 4.35 & 33.33 \\
Japanese & 69 & 0.00 & 5.80 & 4.35 & 2.90 & 2.90 \\
Korean & 69 & 5.80 & 8.70 & 10.14 & 4.35 & 8.70 \\
Portuguese & 69 & 7.25 & 20.29 & 23.19 & 8.70 & 20.29 \\
Spanish & 69 & 1.45 & 15.94 & 21.74 & 1.45 & 20.29 \\
Swahili & 69 & 1.45 & 13.04 & 17.39 & 5.80 & 18.84 \\
Wolof & 69 & 5.80 & 11.59 & 14.49 & 4.35 & 5.80 \\
Yoruba & 69 & 2.90 & 11.59 & 7.25 & 0.00 & 8.70 \\
Zulu & 69 & 1.45 & 1.45 & 4.35 & 2.90 & 0.00 \\
\midrule
\textbf{Total} & 830 & 5.18 & 16.51 & 17.95 & 5.42 & 15.90 \\
\bottomrule
\end{tabular}
}
\caption{Per-language tool-based accuracy for \textsc{Qwen3.6-35B-A3B}. All values are percentages.}
\label{tab:perlang-qwen-tool}
\end{table}

\begin{table}[!htbp]
\centering
\scriptsize
\setlength{\tabcolsep}{2.8pt}
\resizebox{\columnwidth}{!}{
\begin{tabular}{@{}lrccccc@{}}
\toprule
Language & $N$ & BM25 & Q3-4B & Q3-8B & E5 & Arctic \\
\midrule
Chinese & 70 & 0.71 & 20.99 & 22.75 & 0.82 & 18.16 \\
English & 70 & 28.38 & 37.31 & 48.93 & 25.75 & 41.18 \\
French & 69 & 6.80 & 26.41 & 29.59 & 5.39 & 24.35 \\
German & 69 & 7.69 & 49.43 & 36.92 & 2.21 & 34.51 \\
Japanese & 69 & 1.57 & 9.28 & 10.03 & 1.10 & 12.78 \\
Korean & 69 & 0.95 & 19.23 & 17.68 & 1.14 & 15.95 \\
Portuguese & 69 & 2.12 & 29.06 & 24.54 & 3.74 & 22.32 \\
Spanish & 69 & 3.35 & 28.85 & 34.73 & 1.92 & 34.04 \\
Swahili & 69 & 3.50 & 18.37 & 20.85 & 2.18 & 14.12 \\
Wolof & 69 & 2.77 & 18.87 & 16.09 & 2.77 & 7.29 \\
Yoruba & 69 & 4.46 & 12.01 & 13.99 & 3.76 & 15.98 \\
Zulu & 69 & 4.48 & 7.49 & 10.99 & 1.80 & 4.13 \\
\midrule
\textbf{Total} & 830 & 5.59 & 23.12 & 23.95 & 4.40 & 20.42 \\
\bottomrule
\end{tabular}
}
\caption{Per-language evidence recall for \textsc{GPT-OSS-20B}. All values are percentages.}
\label{tab:perlang-gptoss-recall}
\end{table}

\begin{table}[!htbp]
\centering
\scriptsize
\setlength{\tabcolsep}{2.8pt}
\resizebox{\columnwidth}{!}{
\begin{tabular}{@{}lrccccc@{}}
\toprule
Language & $N$ & BM25 & Q3-4B & Q3-8B & E5 & Arctic \\
\midrule
Chinese & 70 & 0.93 & 28.74 & 27.32 & 1.54 & 21.75 \\
English & 70 & 39.85 & 50.82 & 49.60 & 34.31 & 45.33 \\
French & 69 & 11.41 & 34.28 & 31.12 & 5.58 & 28.06 \\
German & 69 & 7.70 & 48.48 & 45.79 & 3.27 & 36.86 \\
Japanese & 69 & 1.83 & 13.14 & 10.78 & 2.43 & 17.60 \\
Korean & 69 & 0.57 & 20.15 & 22.32 & 1.49 & 17.34 \\
Portuguese & 69 & 3.63 & 38.70 & 31.40 & 6.00 & 31.96 \\
Spanish & 69 & 8.10 & 36.03 & 45.56 & 7.09 & 39.05 \\
Swahili & 69 & 6.14 & 18.39 & 31.20 & 3.94 & 21.40 \\
Wolof & 69 & 7.61 & 24.30 & 17.86 & 6.71 & 10.72 \\
Yoruba & 69 & 6.97 & 14.27 & 16.96 & 2.44 & 15.52 \\
Zulu & 69 & 3.76 & 14.39 & 16.05 & 4.04 & 7.11 \\
\midrule
\textbf{Total} & 830 & 8.24 & 28.50 & 28.85 & 6.60 & 24.41 \\
\bottomrule
\end{tabular}
}
\caption{Per-language evidence recall for \textsc{GPT-OSS-120B}. All values are percentages.}
\label{tab:perlang-gptoss120b-recall}
\end{table}

\begin{table}[!htbp]
\centering
\scriptsize
\setlength{\tabcolsep}{2.8pt}
\resizebox{\columnwidth}{!}{
\begin{tabular}{@{}lrccccc@{}}
\toprule
Language & $N$ & BM25 & Q3-4B & Q3-8B & E5 & Arctic \\
\midrule
Chinese & 70 & 0.57 & 24.21 & 27.92 & 0.29 & 17.23 \\
English & 70 & 30.74 & 49.38 & 49.20 & 26.56 & 44.41 \\
French & 69 & 5.18 & 27.47 & 35.15 & 2.68 & 27.47 \\
German & 69 & 5.86 & 36.73 & 37.00 & 1.71 & 28.58 \\
Japanese & 69 & 1.90 & 13.54 & 12.29 & 1.59 & 8.49 \\
Korean & 69 & 0.24 & 21.95 & 17.71 & 0.74 & 16.57 \\
Portuguese & 69 & 3.90 & 28.09 & 29.43 & 5.41 & 23.62 \\
Spanish & 69 & 3.61 & 33.65 & 39.86 & 2.12 & 33.91 \\
Swahili & 69 & 3.60 & 18.40 & 23.70 & 2.64 & 19.52 \\
Wolof & 69 & 6.23 & 16.61 & 20.50 & 5.16 & 8.66 \\
Yoruba & 69 & 5.79 & 12.33 & 15.39 & 2.25 & 13.22 \\
Zulu & 69 & 5.28 & 10.52 & 12.41 & 3.42 & 4.18 \\
\midrule
\textbf{Total} & 830 & 6.10 & 24.44 & 26.74 & 4.57 & 20.51 \\
\bottomrule
\end{tabular}
}
\caption{Per-language evidence recall for \textsc{Qwen3.6-35B-A3B}. All values are percentages.}
\label{tab:perlang-qwen-recall}
\end{table}

\begin{table}[!htbp]
\centering
\scriptsize
\setlength{\tabcolsep}{3.5pt}
\resizebox{\columnwidth}{!}{
\begin{tabular}{@{}lrrrrr@{}}
\toprule
Language & $N$ & BM25 Acc. & BM25 Rec. & Q3-8B Acc. & Q3-8B Rec. \\
\midrule
Chinese & 70 & 10.00 & 3.40 & 51.43 & 56.94 \\
English & 70 & 65.71 & 69.28 & 71.43 & 75.40 \\
French & 69 & 13.04 & 19.09 & 50.72 & 58.59 \\
German & 69 & 21.74 & 21.66 & 57.97 & 70.51 \\
Japanese & 69 & 10.14 & 6.15 & 20.29 & 40.13 \\
Korean & 69 & 4.35 & 0.65 & 42.03 & 47.04 \\
Portuguese & 69 & 14.49 & 12.62 & 56.52 & 60.49 \\
Spanish & 69 & 10.14 & 11.81 & 42.03 & 63.77 \\
Swahili & 69 & 11.59 & 18.57 & 34.78 & 51.30 \\
Wolof & 69 & 15.94 & 16.39 & 31.88 & 42.72 \\
Yoruba & 69 & 17.39 & 24.78 & 24.64 & 44.22 \\
Zulu & 69 & 14.49 & 17.21 & 23.19 & 34.35 \\
\midrule
\textbf{Total} & 830 & 17.47 & 18.51 & 42.29 & 53.82 \\
\bottomrule
\end{tabular}
}
\caption{Per-language tool-based performance for \textsc{DeepSeek-V4-Pro}. Acc. and Rec. denote accuracy and evidence recall; all values are percentages.}
\label{tab:perlang-deepseek-tool}
\end{table}

\begin{table}[!htbp]
\centering
\small
\setlength{\tabcolsep}{5pt}
\begin{tabular}{@{}lrrrr@{}}
\toprule
Language & $N$ & OSS-20B & OSS-120B & Qwen3.6\\
\midrule
Chinese & 70 & 80.00 & 77.14 & 91.43 \\
English & 70 & 91.43 & 92.86 & 92.86 \\
French & 69 & 89.86 & 91.30 & 97.10 \\
German & 69 & 88.41 & 92.75 & 94.20 \\
Japanese & 69 & 57.97 & 75.36 & 73.91 \\
Korean & 69 & 66.67 & 76.81 & 85.51 \\
Portuguese & 69 & 89.86 & 97.10 & 95.65 \\
Spanish & 69 & 86.96 & 92.75 & 89.86 \\
Swahili & 69 & 76.81 & 83.82 & 89.86 \\
Wolof & 69 & 60.87 & 79.71 & 86.96 \\
Yoruba & 69 & 75.36 & 86.96 & 94.20 \\
Zulu & 69 & 66.67 & 76.81 & 78.26 \\
\midrule
\textbf{Total} & 830 & 77.59 & 85.28 & 89.16 \\
\bottomrule
\end{tabular}
\caption{Per-language oracle accuracy in the cross-lingual setting. All values are percentages.}
\label{tab:perlang-oracle-all}
\end{table}

\section{Tongyi-DeepResearch Results}
\label{app:tongyi-results}

We additionally evaluate \textsc{Tongyi-DeepResearch-30B-A3B} \citep{tongyidr}, a deep research agent built on a Qwen3-based MoE architecture. Unlike the other agents in our study, Tongyi uses an in-band ReAct-style tool calling protocol with \texttt{<tool\_call>} XML tags rather than the OpenAI function-calling API. Table~\ref{tab:tongyi-results} reports its performance with BM25 and \textsc{Qwen3-Embedding-8B} across all three corpus conditions. Tongyi's ReAct-style output format does not reliably produce per-query confidence scores despite prompt-level instructions, making the metric calibration error unreliable. Therefore, we exclude it from our main results but put it in the appendix for reference.

\begin{table}[h]
\centering
\small
\setlength{\tabcolsep}{4pt}
\begin{tabular}{llrrr}
\toprule
Retriever & Corpus & Acc. & Ev.Rec. & Search \\
\midrule
\multirow{3}{*}{BM25}
 & Orig. & 22.77 & 34.06 & 22.3 \\
 & Multi. & 9.88 & 17.64 & 35.9 \\
 & Cross. & 8.80 & 17.01 & 35.6 \\
\midrule
\multirow{3}{*}{Q3-8B}
 & Orig. & 39.64 & 58.12 & 27.7 \\
 & Multi. & 26.14 & 48.18 & 34.4 \\
 & Cross. & 25.06 & 44.97 & 34.5 \\
\bottomrule
\end{tabular}
\caption{\textsc{Tongyi-DeepResearch-30B-A3B} results. Acc.\ and Ev.Rec.\ are percentages; Search is average search calls per query. Q3-8B denotes \textsc{Qwen3-Embedding-8B}. Calibration error is omitted because Tongyi's ReAct-style output format does not reliably produce per-query confidence scores despite prompt-level instructions, making the metric unreliable.}
\label{tab:tongyi-results}
\end{table}

Tongyi achieves 39.64\% accuracy on the original corpus with \textsc{Qwen3-Embedding-8B}, the highest among all agents at comparable parameter counts. Its evidence recall (58.12\%) also exceeds \textsc{GPT-OSS-20B} (42.91\%) and \textsc{Qwen3.6-35B-A3B} (43.14\%). After translation, accuracy drops by 13.50--14.58\,pp with \textsc{Qwen3-Embedding-8B}, a smaller relative degradation than \textsc{GPT-OSS-20B} (20.84--20.96\,pp).

\section{Inference Hyperparameters}
\label{app:inference-hyperparameters}

For each agent we follow the generation configuration recommended by the model release, applied uniformly across all corpus conditions and  evidence languages. \textsc{GPT-OSS-20B} and \textsc{GPT-OSS-120B} are served locally with vLLM in temperature $1.0$, top-$p$ $1.0$.  \textsc{Qwen3.6-35B-A3B} is served locally with vLLM in temperature $0.7$, top-$p$ $0.8$. \textsc{DeepSeek-V4-Pro} is accessed through its official API in default settings(temperature $1.0$, top-$p$ $1.0$). All other generation parameters are left at each model's default value.

\section{License Statement}
\label{app:license}
\paragraph{BrowseComp-Plus.}
Our benchmark, \textsc{XBCP}, is derived from BrowseComp-Plus \citep{chen2025browsecompplusfairtransparentevaluation}, which is released under the MIT License. We use BrowseComp-Plus in accordance with the MIT License terms, retaining the original copyright notice and license text in all derived artifacts.

\paragraph{Models.}
We use the following models under their respective licenses: \textsc{GPT-OSS-20B} \citep{openai2025gptoss120bgptoss20bmodel}, \textsc{GPT-OSS-120B} \citep{openai2025gptoss120bgptoss20bmodel}, \textsc{Qwen3-Embedding-4B}, \textsc{Qwen3-Embedding-8B} \citep{zhang2025qwen3embeddingadvancingtext}, and \textsc{Arctic-Embed-L-2.0} \citep{yu2024arcticembed20multilingualretrieval} are released under the Apache License 2.0; \textsc{Multilingual-E5-Large} \citep{wang2024multilinguale5textembeddings} is released under the MIT License. \textsc{Qwen3.6-35B-A3B} \citep{qwen36_35b_a3b} is released under the Apache License 2.0 and is used locally via vLLM. \textsc{DeepSeek-V4-Pro} \citep{deepseekai2026deepseekv4}, whose model weights are released under the MIT License, is accessed in our experiments through its official API under the DeepSeek Open Platform Terms of Service. \texttt{GPT-5.4} \citep{openai2026gpt54thinking}, a proprietary model accessible only through OpenAI's API, is used solely to generate translations for the \textsc{XBCP} evidence corpora; its outputs are used in accordance with OpenAI's Terms of Use, which grant users ownership of model outputs subject to OpenAI's usage policies. All use of these models is for non-commercial academic research.

\paragraph{Release.}
We will release \textsc{XBCP} under the MIT License, consistent with the license of the underlying BrowseComp-Plus benchmark. The release will include the translated evidence corpora, query–language assignments, and evaluation scripts, with attribution to BrowseComp-Plus and to each model whose outputs contributed to the construction of the benchmark.

\section{GenAI Statement}
\label{app:genai}
We disclose the use of generative AI tools in this work in accordance
with the ACL Policy on the Use of AI Writing Assistance.

\paragraph{AI use in research artifacts.}
Generative AI played a central role in constructing the \textsc{XBCP}  benchmark. Specifically, we used \texttt{GPT-5.4} \citep{openai2026gpt54thinking} as the translation  engine to render the English evidence documents of BrowseComp-Plus
\citep{chen2025browsecompplusfairtransparentevaluation} into the eleven non-English target languages used in our cross-lingual and multilingual corpora. The exact prompt is provided in Appendix~\ref{app:translation-prompt}. Translation quality was assessed through expert human verification on samples of all eleven non-English using the rubric in Appendix~\ref{app:verification-rubrics}; we discuss the implications and limitations of automatic translation in the Limitations section.

\paragraph{AI use in experiments.}
The agents and retrievers evaluated in this work are themselves  LLM-based or neural systems (\textsc{GPT-OSS-20B}, \textsc{GPT-OSS-120B}, \textsc{Qwen3.6-35B-A3B},  \textsc{DeepSeek-V4-Pro}, and four multilingual embedding models).  Their use is the subject of study rather than an auxiliary tool, and  is fully described in Section 4.

\paragraph{AI use in writing.}
We used AI assistants (Claude and ChatGPT) for surface-level writing support, including grammar correction, sentence-level rephrasing for clarity and concision, and LaTeX formatting suggestions. All scientific claims, experimental design choices, analyses, and conclusions are authored and verified by the human authors. AI assistants were not used to generate citations, statistical results, or any factual content reported in this paper.

\paragraph{Responsibility.}
The authors take full responsibility for the content of this paper,  including any text that may have been initially drafted or edited with AI assistance.

\section{Ethics}
\label{app:ethics}
XBCP is a translation-based benchmark for evaluating deep research agents. Translations are produced by GPT-5.4, and despite expert verification on a sample, residual translation artifacts may propagate into low-resource-language evaluation, potentially under- or over-estimating system performance for those languages. XBCP is derived from BrowseComp-Plus, which is built from publicly available web documents. We do not collect new personal data from individuals. The benchmark therefore inherits the question scope of BrowseComp-Plus and is intended for research evaluation, not for deployment-grade safety claims.

Expert bilingual annotators were recruited through commercial language-service companies. They were compensated according to standard professional translation-evaluation rates.
\end{document}